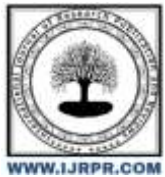

# International Journal of Research Publication and Reviews

# Development of a WAZOBIA-Named Entity Recognition System

## S.E Emedem [1]*, I.E Onyenwe [1], E. G Onyedinma [1]

[1]Department of Computer Science, Nnamdi Azikiwe University, Awka, Anambra State, Nigeria
se.emedem@stu.unizik.edu.ng, ie.onyenwe@unizik.edu.ng, eg.osita@unizik.edu.ng

### ABSTRACT

Named Entity Recognition NER is very crucial for various natural language processing applications, including information extraction, machine translation, and sentiment analysis. Despite the ever-increasing interest in African languages within computational linguistics, existing NER systems focus mainly on English, European, and a few other global languages, leaving a significant gap for under-resourced languages. This research presents the development of a WAZOBIA-NER system tailored for the three most prominent Nigerian languages: Hausa, Yoruba, and Igbo. This research begins with a comprehensive compilation of annotated datasets for each language, addressing data scarcity and linguistic diversity challenges. Exploring the state-of-the-art machine learning technique, Conditional Random Fields (CRF) and deep learning models such as Bidirectional Long Short-Term Memory (BiLSTM), Bidirectional Encoder Representation from Transformers (Bert) and fine-tune with a Recurrent Neural Network (RNN), the study evaluates the effectiveness of these approaches in recognizing three entities: persons, organizations, and locations. The system utilizes optical character recognition (OCR) technology to convert textual images into machine-readable text, thereby enabling the Wazobia system to accept both input text and textual images for extraction purposes. The system achieved a performance of 0.9511 in precision, 0.9400 in recall, 0.9564 in F1-score, and 0.9301 in accuracy. The model's evaluation was conducted across three languages, with precision, recall, F1-score, and accuracy as key assessment metrics. The Wazobia-NER system demonstrates that it is feasible to build robust NER tools for under-resourced African languages using current NLP frameworks and transfer learning. This work lays the foundation for further research and practical applications in NLP, such as machine translation, digital assistants, and other multilingual NLP and NER tasks in an African context, and incorporates more domain-specific data (e.g., legal, health, education, Agriculture).

## 1. Introduction

Named entity recognition (NER) is a method of automatically extracting the most important parts of a text, such as names, locations, dates, times, product names, amounts of money, and more. NER is the process of finding and labeling significant nouns and phrases in a text. Names, places, organizations, dates, amounts, and percentages are all examples of named entities because they serve as the text's focal points. In Natural Language Processing, the term "Named Entity" was coined from the sixth iteration of the Message Understanding Conference (MUC-6) (Grishman & Sundheim, 1996). Information about business and even defense-related activities was extracted from unstructured text like journal and newspaper articles using heuristics and handcrafted rules, this was the primary focus of MUC at the time. Participants often reflect on how important it was to keep track of things like names, locations, companies, times, and amounts of money throughout the activity. In order to better understand the text's context and facilitate a wide range of downstream applications, NER primarily aims to extract and categorize these entities into predefined classes. Over 7,000 languages are currently in use across the globe, but only a tiny percentage of those are considered to be "well-resourced languages." The number of languages spoken on the African continent varies widely but is estimated to be between 1,500 and 2,000 (Sawe, 2018). Different communities even within a single country like Nigeria speak a variety of indigenous languages. Nigerians speak over 522 distinct languages, the most widely spoken of which are Hausa, Yoruba, and Igbo (Akindele, Olatundun, & Akano, 2022), with English serving as the country's Lingua franca. The Igbo language is spoken by 18% of Nigeria's population, Hausa by an average of 25% of Nigeria's population, and about 21% of Nigerians speak the Yoruba language (Statista, 2024), other African countries like Togo, Republic of Benin, Ghana, Sudan, Sierra-Leone, Cote D'Ivoire among others outside Africa, also have speakers of these languages amongst them. Over the years, a good proportion of NER projects have been devoted to the English language and some prominent key languages, due to their importance as a dominant language that is used both internationally and English locally here in Nigeria for Business, Politics, importation, and exportation, science, Information technology, and Entertainment. To this effect, there have been lots of limitations in the field of NLP, as well as NER due to the fact that African languages are under-resourced with little linguistics corpus (Adelani, Hedderich, Zhu, Berg & Klakow, 2020). For instance, most science research works are all carried out in English in almost all countries, due to this, there has been no NER research for under-resourced languages in areas such as drugs industries, chemical named entities, bioinformatics or biomedical NER (Shaalan, 2014). Wazobia is a term used for the combination of three words from the Hausa, Yoruba, and Igbo languages. The word "Wa" comes from the Yoruba language and means "Come," "Zo" comes from the Hausa language and also means "Come," and "Bia" comes from the Igbo language which also means "Come." These words when put together, form the phrase "Wazobia" meaning Come (Fasan, 2015). What the Wazobia-NER system does is to distinguish between entities in Hausa, Igbo, and Yoruba languages, identify the entities, and effectively categorize them accordingly. The Wazobia-NER system was trained on a massive dataset of text that includes all three languages for it to correctly recognize and categorize entities. The Wazobia NER system focuses primarily on identifying and extracting three entities: individual names, locations, and organizations as entities from a text document and from textual image documents. A well-performing Wazobia-NER system will bring a positive increase in the field of NER and NLP, as there are very few systems currently in existence for these languages. The goal of developing a Wazobia-NER System is to facilitate the rapid and precise recognition and classification of named entities from text written in Hausa, Yoruba, and Igbo. Natural language processing relies heavily on the recognition and extraction of named entities because of their significance and practical applications such as question-answering systems, chatbots and virtual assistants, Information retrieval, etc. Wazobia-NER would facilitate the greater use of under-resourced languages, increase opportunities for multilingual communication, cultural preservation, and cooperation, Tourist to understand local names, places, and cultural references, digital inclusion by making digital content more accessible and useable, customer services for speakers, and help develop new tools and resources for these languages.

## 2. Related Works

(Chukwuneke, Rayson, Ezeani, El-Haj, Asogwa, Okpalla & Mbonu, 2023) utilized an existing English NER tool in a cross-language projection method to create a mapping dictionary of entities from a source language to a target language using a parallel English-Igbo corpus thereby addressing the challenges for low-resource languages which one has been the scarcity of annotated training data, particularly for tasks like named entity recognition (NER). This dictionary, after being manually verified by human annotators, was employed to automatically generate and format a NER training dataset from the Igbo monolingual corpus, thereby significantly reducing the annotation time for the Igbo NER task. Incorporating this generated dataset into the training process yielded improved performance compared to previous efforts.

(Abdulmumin & Galadanci, 2019) introduced hauWE (Hausa Words Embedding) models, developed using Word2Vec's Continuous Bag of Words (CBoW) and Skip Gram (SG) models. Word embeddings, which are distributed word vector representations, play a crucial role in various natural language processing (NLP) tasks such as machine translation, sentiment analysis, word analogy, named entity recognition, and word similarity. However, the only existing word vectors for the Hausa language are those trained using fastText, comprising a limited number of word vectors. These new models are significantly larger and more effective than the previous fastText model, enhancing their utility in NLP tasks.

In this study, (Ifeoluwa, Neubig, Ruder, Rijhwani, Nakatumba-Nabende, Ogundepo & Klakow, 2022) alighted how African languages are underrepresented in NLP research and development and the challenges of these shortcomings. The researchers progressed towards finding solutions for these challenges by focusing on the task of named entity recognition (NER) thereby creating the largest human-annotated NER dataset for 20 African languages and providing strong baseline results on the corpus by fine-tuning multilingualPLMs on in-language NER and multilingual datasets. The behaviour of the state-of-the-art cross-lingual transfer methods in an Africa-centric setting demonstrated that the choice of source language significantly affects performance. The analysis and cross-lingual transfer in an Africa-centric showed that choosing the best transfer language improves zero-shot F1 scores by an average of 14 points across 20 languages compared to using English.

(Adelani, Hedderich, Zhu, Berg & Klakow, 2020) focus on NER for Hausa and Yoruba, evaluating different embedding approaches and demonstrating that distant supervision can significantly improve classifier performance in realistic low-resource settings, more than doubling its effectiveness. The development of natural language processing tools like named entity recognition (NER) for many languages in developing countries has been hampered by a lack of labelled training data. Techniques such as distant and weak supervision can create labelled data semi-automatically, while noise-handling methods can mitigate errors in automatic annotation. Pretrained word embeddings are crucial for neural NER classifiers, and the advent of complex contextual embeddings presents a trade-off between model size and performance. Although these techniques are effective in high-resource settings, their performance in low-resource scenarios needs exploration.

(Al-Smadi, Al-Zboon, Jararweh & Juola, 2020) examined the effects of implementing a transfer learning model, specifically the Pooled-GRU model, in comparison to their prior Bi-LSTM CRF model, with the aim of enhancing the Arabic-named entity recognition. Both models went into testing using the WikiFANEGold dataset, and the findings indicated significant performance improvements for our newly proposed model, with an approximate 17% increase in F-measure. In contrast, the authors achieved significantly higher levels of accuracy in their newly proposed models, with a recorded rate of 91.20%, compared to the 75.73% accuracy rate seen in their earlier research. The findings of their research demonstrate notable outcomes and superior performance in comparison to the emerging ANER systems.

(Omisore 2023) introduced a named entity recognition approach specifically for Yoruba, utilizing Long Short-Term Memory (LSTM) networks. Named entity recognition is crucial for understanding narratives, especially in low-resourced languages like Yoruba. The LSTM-based model is designed to identify both named entities and event entities across different event types. By integrating a Time Distributed Output layer, the model ensures precise temporal alignment and parallelization for accurate event trigger identification. Experiments on a Yoruba event ex-traction dataset reveal the model's effectiveness, achieving an F1-score of 0.8967%, precision of 0.9391%, and recall of 0.8579%, highlighting its robustness and adaptability in the Yoruba linguistic context.

(Afeez, Ibrahim, Bolaji & Adedayo, 2024) developed three Nigerian Word character recognition using ResNet-50 an architecture of Convolutional Neural Networks. The conversion of typed image text into machine-readable format has become a necessity as the majority of typing and communication is being done via smartphones. Moreover, the understanding of these texts like translation warrants it being first recognized. 100 different words per language were acquired over the internet and converted to images using Python script. These images were pre-processed using normalization, transformation, rescaling, and feature extracted using a Histogram of Oriented Gradient. The extracted feature was then fed into ResNet-50 for recognition and classification with an accuracy of 77% while Yoruba gave the best weighted precision of 89%. The evaluated result was further compared with some machine learning algorithms like K-nearest Neighbour and Support Vector Machine and the system using ResNet-50 outperformed the earlier mentioned model.

 (Shahshahani, Mohseni, Shakery, & Faili, 2018) developed a novel annotated dataset for Persian named-entity recognition, which they named Peyma. The Peyma dataset comprises a total of 7,145 sentences, consisting of 302,530 tokens. Among these tokens, 41,148 are annotated with entity tags. These annotations were acquired from a corpus of 709 documents.

 (Ray Chawdhary, Das, Dozier & Seals, 2023) developed a sentiment analysis (for positive and negative as well as neutral) system for tweets from low-resource African languages: Igbo and Hausa, the researchers worked on the first thorough compilation of AfriSenti-SemEval 2023 Shared Task 12 Twitter datasets that are human-annotated for the most widely spoken languages in Nigeria such as Hausa and Igbo. The researchers trained the modern pre-trained language model AfriBERTa large on the AfriSenti-SemEval Shared Task 12 Twitter dataset to create sentiment classification, the results demonstrate that the model trained on AfriSenti-SemEval Shared Task 12 datasets and produced with an F1 score of 80.85% for Hausa and 80.82% for Igbo languages on the sentiment analysis test.

 (Adegunlehin, Asahiah & Onifade, 2019) investigated the feature characteristics of Yoruba NER, the researchers explored two feature characteristics - Surrounding words and the Part-of-Speech (PoS) tag of Named Entities as it can be applied to the recognition of Yoruba Named Entities while focusing on only three named entities namely Person, Location and Time entities in a Yoruba text. The classification model was formulated using Conditional Random Fields (CRFs) classifier and a feature investigation experiment was carried on the result of the classification. The result of the experiment showed the impact of each of the features: PoS tag as a feature in Yoruba NER helps to identify a word as a Named Entity while information about the Surrounding words helps to classify a word into its Named Entity class.

## 3. Materials and Methods

**Architectural Design** represents the **structural components of the system's** overall organization and how they interact.

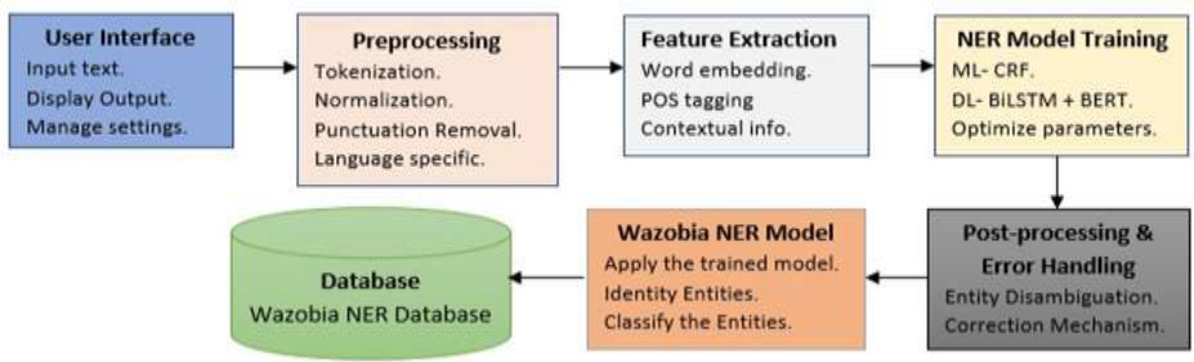

Figure 1. Architectural Design of the System

**User Interface (UI):** This is where users interact with the system. The UI enables users to input text in Hausa, Yoruba, or Igbo, view the extracted entities (i.e., names, locations, and organizations), and manage settings, such as selecting language preferences.

**Pre-processing:** This step prepares the input text, including tokenization, which splits sentences into words, Normalization, which converts text to lowercase and standardizes spelling, Punctuation Removal, which removes unnecessary symbols that may affect entity recognition, and Language-Specific Processing, which adapts the pre-processing based on Yoruba, Igbo, or Hausa grammar rules.

**Feature Extraction**: The module extracts important linguistic and contextual features from the text. The Key tasks include Word Embedding for Transforming words into vector representations using Word2Vec, POS Tagging (Part-of-Speech Tagging) for identifying word categories (e.g., nouns, verbs, adjectives), and Contextual Information for extracting surrounding words to improve entity recognition.

**NER Model Training**: Here, the model is trained to recognize named entities from text. Machine Learning Conditional Random Fields (CRF) were used for classification, Deep Learning advanced models like Bidirectional LSTM (BiLSTM), and BERT-RNN for better multilingual entity recognition support, and Parameter Optimization for Fine-tuning the models to improve accuracy.

**Post-processing & Error Handling:** The system applies **Entity Disambiguation** for resolving ambiguous entities (e.g., "Lagos" as a location and "Lagos Food Bank" as an organization) and correction **Mechanisms** for fixing errors due to spelling variations in Hausa, Yoruba, and Igbo.

**Wazobia-NER Model:** This is the final trained model that is used to identify and classify entities in new text inputs. It applies the trained model to process new text data identify the entities, and classify the entities based on the recognized categories.

**Database:** The extracted entities are stored in a structured database for future use and to improve future training by refining the dataset.

*3.1 Dataset and Preparation*

Developing a Wazobia-NER system involves preparing relevant datasets to train and fine-tune the NER model.

**Data Collection:** A diverse dataset of articles was selected from a collection of online news articles, journals, and academic papers written in Hausa, Igbo, and Yoruba languages. A total of 20 articles were selected for each of the languages used. These datasets collected contained the specific entities needed and were used to train the Wazobia NER model to recognize and classify entities from text effectively.

**Data Annotation:** Annotated data with ground truth labels (entities and their types) was used to train the NER model. Pre-trained multilingual model (like **BERT) was used to suggest labels automatically, while the** Bi-LSTM model was used for training the model. **The** annotated dataset was integrated with Optical Character Recognition (OCR) and made textual images machine-readable.

**Fine-tuning data:** After pre-training the model with the Wazobia datasets, a smaller amount of annotated (labelled) datasets is fed to the model. This helped the model to adapt to the generalizing of entities so as not to forget the knowledge already gathered and prevent overfitting or underfitting.

**Training-validation-testing split:** The datasets collected were divided into training and testing datasets to a ratio of 80:20. The Testing dataset was further divided into Validation and Testing datasets. The training dataset was used to train the model, the validation dataset was used to fine-tune hyper-parameters, and for the testing dataset, an unseen dataset was used to evaluate the model's overall performance.

**Evaluation Metrics:** Metrics like precision, recall, and F1-score were used to assess the model's performance. Precision measures the accuracy of the model in identifying entities, while recall measures the completeness.

TABLE I: Classification Evaluation of the Wazobia-NER system on test data

| Epoch | Training Loss | Validation Loss | Precision | Recall | F1-score | Accuracy |
|---|---|---|---|---|---|---|
| 1 | 1.791200 | 1.850040 | 0.941902 | 0.960161 | 0.961616 | 0.922514 |
| 2 | 1.765200 | 1.783243 | 0.918861 | 0.973340 | 0.972459 | 0.963589 |
| 3 | 1.753800 | 1.776508 | 0.926274 | 0.963694 | 0.930162 | 0.967110 |
| 4 | 1.762600 | 1.774690 | 0.975257 | 0.935221 | 0.959462 | 0.962875 |
| 5 | 1.764400 | 1.772275 | 0.951086 | 0.939985 | 0.956417 | 0.930136 |

Table 1 shows the evaluation of the model during the training process for the model including the "Training loss", "Validation loss", "Precision", "Recall", "F1-score" and "Accuracy" at each Epoch/iteration to measure and check for the accuracy of the model and if the model needs further fine-tuning to get better and more efficient model.

An Epoch is reached when the Model has seen all the datasets for training once. For a Model to efficiently generalize well to unseen data and entities, it should be trained for multiple Epochs, just as humans learn from consistent practices, so do machines and models amplify over continuous practices. For the Wazobia-NER model, the model was trained for 50 Epochs. The Training Loss is the error calculated on the training dataset after each Epoch, and it tells how well the model is learning from the training dataset. The Validation Loss is the error calculated on the Validation dataset, and it tells how well the model is generalizing to unseen data.

Precision measures the number of entities the model identified as entities that are actual entities. It checks how many entities the Wazobia-NER model identified are correct. Recall measures the actual relevant entities the model identifies as entities. It checks for the number of actual entities the model successfully found from the text. F1-Score measures the harmonic mean between the Precision and Recall. It ensures that both the incorrectly identified entities and the missed entities are all accounted for. Accuracy measures the proportion of correctly identified entities by the model.

*3.2 Experimental Tools*

The system was developed using various technologies and tools such as Python, Flask framework, Tesseract, NLTK, Spacy, SQLite, and Bootstrap. Python v3.8 is an open-source programming language that offers extensive libraries and frameworks such as Flask, PyTorch, NLTK, Spacy, used for developing interactive web applications both for front-end and back-end, to build and train the model, and Scikit-learn for evaluating the model's performance. Python's syntax is clear, simple to learn, use, read, and reusable. SQLite is a relational database management system that is used to interact with Python for storing the annotated data and model configurations. Tesseract is an OCR tool, preferably suitable for languages to convert textual images into machine-readable text. Bootstrap provides a collection of pre-designed HTML, CSS, and JavaScript tools used to create highly efficient, responsive, and visually stunning websites and web applications.

## 4. Results and Discussion

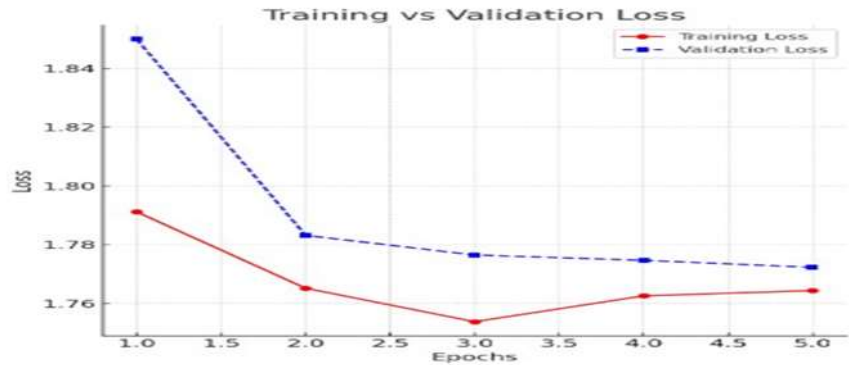

Figure 2. Line Chart of Training Loss and Validation Loss.

Figure 3 represents the Line chart of the training and validation loss of the model over 5 epochs. Initially, both are high but decrease after the first epoch, showing improvement. Training loss shows steady declines, indicating that the model is learning well, while the validation loss shows that the model is generalizing well to unseen data.

From Figure 4, it is shown from the Graph below that Precision peaks at Epoch 4, meaning the model had fewer incorrect label entities, and Recall peaks at Epoch 2, meaning the model had its highest ability to detect entities. F1-score is highest at Epoch 2, meaning the best balance between precision and recall. Accuracy is highest at Epoch 3 for the total sample.

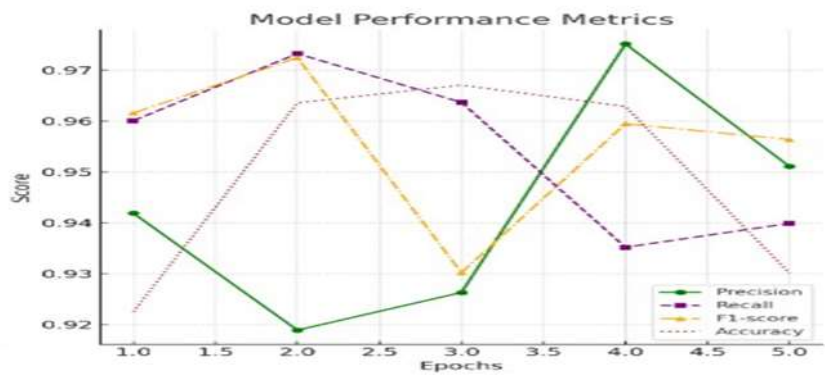

Figure 3. Line Chart for Precision, Recall, F1-Score, and Accuracy.

## 5. Conclusion

The study successfully

developed a high-performing NER system for Igbo, Hausa, and Yoruba languages. By leveraging advanced deep learning techniques, specifically the integration of BERT and RNN, we achieved substantial improvements in entity recognition accuracy. The web application further demonstrated the system's practical utility, offering a user-friendly platform for performing NER tasks. The study underscores the importance of developing NLP tools for underrepresented languages, contributing to the digital inclusion of these linguistic communities. The results indicate that sophisticated models like BERT can be effectively adapted to diverse linguistic contexts, paving the way for future research and development in this area.